\documentclass[wcp]{jmlr}

\usepackage{longtable}
\pdfoutput=1

\usepackage{booktabs} 
\usepackage{algorithmic}
\usepackage{enumerate}
\usepackage{tabu}  
\usepackage{multirow}
\usepackage{graphicx}
\usepackage{caption}
\usepackage{setspace}
\usepackage{graphicx}
\usepackage{algorithm2e}

\jmlrvolume{}
\jmlryear{2019}
\jmlrworkshop{ORSUM 2019}

\title{Optimal Delivery with Budget Constraint in E-Commerce Advertising}

\author{\Name{Chao Wei}, \Name{Weiru Zhang}\footnotemark[1]
\Email{weichao.wc,weiru.zwr@alibaba-inc.com}\\
\Name{Shengjie Sun}\Email{shengjie.ssj@alibaba-inc.com}\\
\Name{Fei Li}\Email{faker.lf@alibaba-inc.com}\\
\Name{Xiaonan Meng}\Email{xiaonan.mengxn@alibaba-inc.com}\\
\Name{Yi Hu}\Email{erwin.huy@alibaba-inc.com}\\
\Name{Kuang-chih Lee}\Email{kuang-chih.lee@alibaba-inc.com}\\
\Name{Hao Wang}\Email{longran.wh@alibaba-inc.com}\\
\addr Alibaba Group\\
\addr Hangzhou, China
}

\begin{document}

\maketitle

\footnotetext[1]{Both authors contributed equally to this research.}

\begin{abstract}
Online advertising in E-commerce platforms provides sellers an opportunity to achieve potential audiences with different target goals. Ad serving systems (like display and search advertising systems) that assign ads to pages should satisfy objectives such as plenty of audience for branding advertisers, clicks or conversions for performance-based advertisers, at the same time try to maximize overall revenue of the platform. In this paper, we propose an approach based on linear programming subjects to constraints in order to optimize the revenue and improve different performance goals simultaneously. We have validated our algorithm by implementing an offline simulation system in Alibaba E-commerce platform and running the auctions from online requests which takes system performance, ranking and pricing schemas into account. We have also compared our algorithm with related work, and the results show that our algorithm can effectively improve campaign performance and revenue of the platform.
\end{abstract}
\begin{keywords}
ad allocation, e-commerce advertising, simulation system
\end{keywords}

\section{Introduction}
E-commerce platforms such as Amazon, eBay and Alibaba run hundreds of millions of auctions to sell advertising opportunities. Online advertising plays a crucial role in connecting advertisers and audiences, and generates tremendous value for E-commerce platforms. There are different types of online advertising which includes performance-based advertising, branding guaranteed advertising. We focus on the performance-based advertising in the paper. In this marketplace, advertisers can specify the maximum daily amount they are willing to pay and get the audience through various pages in E-commerce platform. The objectives of performance-based advertisers are usually to spend out the budget to maximize the performance goals (e.g., clicks, conversions as many as possible). Meanwhile, the ad serving system is optimizing revenue on behalf of the platform.

One of the central issues for the ad serving system of E-commerce platform is matching ads to requests with these objectives above, which can be formulated as a constrained optimization problem. There are many challenges to achieve all the objectives simultaneously in a complex competition environment. Each individual campaign has its own budget and performance goal, and there are hundreds of thousands of campaigns which compete with each other to acquire inventory in the marketplace. These varieties make the optimization extremely difficult. 

In this paper, we present our work on optimal delivery in E-commerce platform, which can be formulated as a constrained optimization problem that maximizes specified goals and subjects to budget constraints. Our contribution can be summarized as follows:
\begin{itemize}
	\item We propose an approach based on linear programming to optimize overall revenue of the platform and improve different performance goals in campaign level simultaneously. Compared with previous work, our approach can be run over all traffic and implemented in both display and search advertising. 
	\item We design a simulation system to evaluate the results of different allocation plans by replaying auction from online requests. The proposed approach has been implemented in conjunction with pricing and ranking schemes in Alibaba E-commerce platform, and the results show that it can effectively improve both revenue and various performance goals.
\end{itemize}

\section{Related Work}
Most existing work related to optimizing budget constrained spend focused on bid modification and allocation. Allocation treats bids as fixed, and allows only decisions about whether the advertiser should participate in the auctions while bid modification is in a setting where bids can be changed \cite{karande2013optimizing}. \cite{bhalgat2012online,chen2011real,zhang2014optimal,devanur2011near} formulated display ad allocation problem by combining budget and bid optimization. \cite{zhu2009revenue,agarwal2014budget} used probabilistic throttling to achieve budget control in sponsored search advertising. \cite{xu2015smart} suggested using allocation or throttling to directly influence budget spending while bid optimization changes the win-rate to control. We also choose allocation strategy for consideration that advertisers in E-commerce platform usually set fixed price or price range for each campaign so that the performance of bidding optimization is limited.

There are several papers close to our work. \cite{abrams2007optimal} solved the problem by an optimal algorithm, but it can only run over head queries. Our approach uses request instead of query in order to be implemented in both search and display advertising. \cite{karande2013optimizing} used a non-optimal algorithm and thought it is fairer than optimal algorithm. We still use optimal algorithm because it can avoid unnecessary competitions and get better performance which is proved by the experiments in Section 5. Motivated by \cite{chen2011real} and \cite{xu2015smart}, we define a multi-objective optimization problem, and convert it into a single-objective optimization problem with constraints since there may be multiple Pareto optimal solutions to a multi-objective optimization problem. We try to maximize the overall revenue and improve different performance goals simultaneously based on linear programming. The widely noted work by \cite{mehta2013online} presented the original formulation and there are related work such as \cite{zhang2018whole, chen2012ad, lee2013real, he2013game, chervonenkis2013optimization} extended this framework to solve specific allocation problems such as dynamic ad allocation, video-ad allocation and smooth budget delivery.
  
\section{Problem And Formulation}
\subsection{Preliminaries}
What most ad serving systems do is to let advertisers participate in auctions as many as possible until their budgets have been exhausted, and then make them ineligible for the rest of the day. Obviously, this strategy causes stronger auction competition at the beginning of the day and yields very biased traffic to the advertisers, which is hard to maximize advertisers' performance goals. In this paper, we study an online allocation strategy to achieve different performance goals and maximize revenue simultaneously. Before formulating this problem, we define some basic concepts. Thus we denote by:

When a user browse a page (in the recommender system) or types a keyword query (in the search engine), there is a request sent to the ad serving system. $\Re =\{r_1,r_2,...,r_{M}\}$ is the request stream which arrives over the whole day. 

$O$ is the set of ads. 

$c_k \in C=\{c_1,c_2,...,c_{N}\}$  is a campaign comes with a daily budget $budget_{k}$. Advertisers use campaign as a minimum marketing unit to achieve their specific performance goal. 

We consider auctions of each request, where there are bidders competing for P slots. $L_i=\{o_p:o_p \in O, p=1,2,...,P_i\}$ is an ordered set of ad indices called \textit{bidding landscape} , where $P_i$ is the number of ads in the landscape, and $o_p$ is recalled and ranked by the ad serving system. Each \textit{bidding landscape} $L_i$ can be mapped into a set of slates.

$L_i^j=\{o_l^j:o_l^j \in L_i, l=1,2,...,P_i^j, P_i^j \le P\}$ is slate $j$ for request $i$, where $P$ is the maximum number of slots available for advertising on the page. \textit{slate} can be obtained by deleting ads of $L_i$ (while maintaining the ordering) and then truncating (if necessary) to $P_i^j$ (at most $P_i$) ads. Generally speaking, $P$ is much smaller than $P_i$. \textit{Slate} represents a unique subset of a \textit{bidding landscape}.

These two crucial concepts of \textit{bidding landscape} and  \textit{slate} of ads we define are proposed by \cite{abrams2007optimal}.

$\Omega_k$ refers to the set of slates including campaign $k$.

$x_{ij}$ is a binary variable indicating whether slate $j$ is assigned to request $i$ ($x_{ij}=1$) or not ($x_{ij}=0$).

\subsection{Problem Definition}
Given the above notations, we can formulate the problem mentioned before as a multi-objective linear programming (MOLP) problem as follows:
\begin{equation}
\begin{aligned}
&\max &    & \qquad \qquad \sum_{i,j}rev_{ij}\cdot x_{ij} \\
&\max &              & \sum_{k \in C_1}\sum_{i,j \in \Omega_{k}}ctr_{ijk}\cdot x_{ij} \\
&\max &              &\sum_{k \in C_2}\sum_{i,j \in \Omega_{k}}cvr_{ijk}\cdot x_{ij} \\
&s.t. &\forall k,    & \sum_{(i,j) \in \Omega_{k}}cost_{ijk}\cdot x_{ij}\le budget_{k} \\
&     &\forall i,    & \qquad \qquad \sum_{j}x_{ij} \le 1, \\
&     &\quad \forall i,j,  & \qquad \qquad \qquad x_{ij} \ge 0.
\end{aligned}
\end{equation}

In Eq.(1), $rev_{ij}$ denotes the expected revenue from slate $j$ for request $i$. $cost_{ijk}$ is the expected cost of campaign $k$ when slate $j$ is assigned to request $i$. The first objective is the expected revenue of platform. In real scenarios, advertisers often prioritize their performance objectives for different campaigns. Other objectives of Eq.(1) are the different performance goals in campaign level such as click and conversion. As there usually exist multiple (possibly infinite) Pareto optimal solutions, it's very difficult to solve such a multi-objective optimization problem. Although the MLOP tries to maximize performance of each campaign, our model can not insure that each campaign gets better result theoretically. Accordingly, we combine all the performance goals of campaigns into several performance constraints and convert the original problem with multiple objectives into a single-objective optimization problem with the constraints. Finally, we formulate the problem as a single-objective linear programming (SOLP) problem with the constraints that advertisers bid for more clicks and conversions in campaign level which is suitable for E-commerce scenario. The formula is presented as follows:
\begin{equation}
\begin{aligned}
&\max &             &\sum_{i,j}rev_{ij}\cdot x_{ij}\\
&s.t. &             &\sum_{k \in C_1}\sum_{i,j \in \Omega_{k}}ctr_{ijk}\cdot x_{ij} \ge T_{cy}\\
&     &             &\sum_{k \in C_2}\sum_{i,j \in \Omega_{k}}cvr_{ijk}\cdot x_{ij} \ge T_{vy}\\
&     &\forall k,   &\qquad \sum_{(i,j) \in \Omega_{k}}cost_{ijk}\cdot x_{ij} \le budget_{k} \\
&     &\forall i,   &\qquad \sum_{j}x_{ij} \le 1, \\
&     &\forall i,j, &\qquad x_{ij} \ge 0.
\end{aligned}
\end{equation}

$ctr_{ijk}$ is the estimated click-through rate of ad from campaign $k$ of appearing in slate $j$ for request $i$ while $cvr_{ijk}$ is the estimated conversion rate. These CTRs and CVRs are predicted based on the real log data and will be used in our algorithm and the simulation system described in Section 4.3. $C_1$ and $C_2$ are the sets of campaigns that bid to achieve more clicks and conversions respectively. Thus the problem is defined by grouping campaigns with the same performance goal(ctr or cvr) and setting click constraint $T_{cy}$ and conversion constraint $T_{vy}$. The constraint setting is discussed in Section 5.

The dual of (2) can be written as follows:
\begin{equation}
\begin{aligned}
&\begin{aligned}
&\min  &                &\sum_{k}\alpha_{k}budget_{k}+\sum_{i}\beta_{i} - \gamma T_{cy} - \delta T_{vy}\\
&s.t.  & \forall i,j,   &\qquad \beta_{i} \ \ \ge rev_{ij}-\sum_{k \in \Omega_{(i,j)}}\alpha_{k}cost_{ijk} \\
&      &                &\qquad \qquad +\gamma\sum_{k \in C_1}ctr_{ijk}+\delta\sum_{k \in C_2}cvr_{ijk}
\end{aligned} \\
&\begin{aligned}
\qquad &      & \forall i,     &\qquad \beta_{i}  &\ge \quad 0 \\
&      & \forall k,     &\qquad \alpha_{k} &\ge \quad 0 \\
&      &                &\qquad \gamma     &\ge \quad 0 \\
&      &                &\qquad \delta     &\ge \quad 0
\end{aligned}
\end{aligned}
\end{equation}

The parameters $\alpha_k$, $\beta_i$, $\gamma$ and $\delta$ are the dual variables of constraints in Eq.(2). There are natural interpretations that $\alpha_k$ can be interpreted as the minimum revenue margin required by a campaign, and $\beta_i$ is the shadow price of satisfying an additional request $i$ with a slate. In a way, $\gamma$ and $\delta$ represent the trade-off between revenue and advertisers' performance goals (click and conversion). $\alpha_k$, $\gamma$ and $\delta$ will be used in the online algorithm in Section 3.3. $\Omega_{(i,j)}$ is the set of campaigns in slate $j$ for request $i$.

\subsection{The Real-Time Algorithm}
Inspired by the previous work of \cite{chen2011real,zhang2018whole}, we develop a real-time algorithm based on the complementary slackness theorem to solve the optimization problem with various constraints mentioned above. The algorithm is shown as follows.

\begin{algorithm2e}
\caption{Real-Time Auction-Algorithm}
\label{alg:net}
\KwIn{$\alpha_k, \gamma, \delta$}
\KwOut{$x_{ij}$}
\For{request $i$ from an online stream}{
  $J \leftarrow \emptyset $\;
  $L_i \leftarrow \textit{bidding landscape generated for i}$\;
  \While{$(generate\ L_i^j )and( L_i^j \neq \emptyset)$}{
    $J \leftarrow J \cup L_i^j$
  }
  $j^{'} \leftarrow argmax_{j \in J} (rev_{ij}-\sum_{k \in \Omega_{(i,j)}}\alpha_{k}cost_{ijk} + \gamma\sum_{k \in C_1}ctr_{ijk}+\delta\sum_{k \in C_2}cvr_{ijk})$\;
  \eIf{$(rev_{ij^{'}}-\sum_{k \in \Omega_{(i,j^{'})}}\alpha_{k}cost_{ij^{'}k} + \gamma\sum_{k \in C_1}ctr_{ij^{'}k}+\delta\sum_{k \in C_2}cvr_{ij^{'}k}) \ge 0$}{
    $ x_{ij^{'}} = 1 $\;
    $ x_{ij} = 0, \forall j \not= j^{'} $\;
  }{
    $ x_{ij} = 0, \forall j$\;
    $\textit{no ads displayed}$\;
  }
}
\end{algorithm2e}

For an incoming request \textit{i}, the ad serving system yields the bidding landscape $L_i$ by retrieving and ranking ads. Each slate $j$ can be obtained by deleting ads of $L_i$ (while maintaining the ordering) and then truncating (if necessary) to $P_i^j$ ads. We will detail how to generate the slate candidates efficiently in next section. Then we calculate $rev_{ij}$, $cost_{ijk}$ and $ctr_{ijk}$ for each ($i$,$j$,$k$) which will be introduced in Section 4.1. Finally we compare each slate $j$ for request $i$ with the score of \begin{math} (rev_{ij}-\sum_{k \in \Omega_{(i,j)}}\alpha_{k}cost_{ijk} + \gamma\sum_{k \in C_1}ctr_{ijk}+\delta\sum_{k \in C_2}cvr_{ijk}) \end{math}, and slate $j$ with the maximal value will be chosen.

\cite{chen2011real} have proved that the online algorithm uses complementary slackness theorem to assign slate to request with the highest scores such that the offline optimality can be preserved. The input $\alpha$, $\gamma$ and $\delta$ can be calculated in advance with the real log data by Equation (3).
\section{Implementation}
There are some practical problems that must be considered and overcome in practice. These include metric prediction, performance challenges of online serving system and offline  simulation system.

\subsection{Metric Prediction}
We first introduce $ctr_{ijk}$ which is the estimated click-though rate of ad from campaign $k$ in slate $j$ for request $i$. Each ad has a specific position $p$ in slate $j$ which means that $ctr_{ijk}$ can be denoted as $ctr_{ij_pk}$. In consideration of position bias, we estimate $ctr_{ij_pk}$ with \textit{Examination Hypothesis}. The Hypothesis \cite{craswell2008experimental,dupret2008user,chapelle2009dynamic} says that users are more likely to click the first rank item and less likely to look at items in lower ranks, which suggests that each rank has a certain probability of being examined. We have $ctr_{ij_pk}$ with denoting this probability as $P(e|p)$:

\begin{equation}
\begin{split}
ctr_{ijk} = ctr_{ij_pk} = P(e|p)\cdot P(c|i,k)
\end{split}
\end{equation}

where $P(c|i,k)$ is the click-through rate of ad from campaign $k$ assigned to request $i$. Assuming under the $PPC$ (pay per click) advertising model, similarly, we have the cost of campaign $k$ assigned to request $i$ in slate $j$:

\begin{equation}
\begin{split}
cost_{ijk} = cost_{ij_pk} = ctr_{ij_pk} * clickprice_{ij_pk}
\end{split}
\end{equation}

The real cost that advertiser should pay when yielding a click is determined by the $GSP$ (generalized second price) mechanism \cite{edelman2007internet} in our ad serving system. When the slate $j$ is fixed, we can easily calculate the click price and the expected cost of campaign $k$ assigned to request $i$.

Assuming the independence of CTRs on the same page, we can express $rev_{ij}$ of request $i$ and slate $j$ in the algorithm as the sum of all the individual expected cost per exposure (\textit{sum method}):

\begin{equation} 
rev_{ij} = \sum_{p=1}^{P}rpm_{ij_pk} = \sum_{p=1}^{P} ctr_{ij_pk}\cdot clickprice_{ij_pk}
\end{equation}

This may not always be a valid assumption, while it is easy to implement and efficient. \cite{zhang2018whole} proposed an \textit{interactive method} that try to estimate $rev_{ij}$ more accurately. The method takes into account the interactive influences across ads on the same page. However, due to the huge number of slates generated from bidding landscape, the \textit{interactive method} brings a great challenge of system performance what we will discuss in the next subsection.

\subsection{Slate Generation}
The intuitive solution of generating slates is to delete some ads in the bidding landscape $L_i$ which will yield at most $(\tbinom{P}{P_i} + P)$ of slates. The time complexity of this solution is $\mathcal{O}(\tbinom{P}{P_i} + P)$ which brings a great challenge of the offline training and online serving. Here we propose a simple method to solve the problem based on the \textit{sum method} mentioned in Section 4.1.

In Algorithm 1, we choose the best slate with the highest score of: 
\begin{equation} 
rev_{ij}-\sum_{k \in \Omega_{(i,j)}}\alpha_{k}cost_{ijk} + \gamma\sum_{k \in C_1}ctr_{ijk}+\delta\sum_{k \in C_2}cvr_{ijk}
\end{equation}

which means that the score of every slate must be calculated. We use \textit{sum method} to approximate Eq.(7) as follows:
\begin{equation} 
\begin{split}
\sum_{p=1}^{P}(rpm_{ij_pk} - \alpha_kcost_{ij_pk}+\gamma I_{C_1}(k) + \delta I_{C_2}(k)) \\
I_{C_1}(k)=
\begin{cases}
ctr_{ij_pk},  & \text{if $k \in C_1$} \\
0, & \text{else}
\end{cases}\\
I_{C_2}(k)=
\begin{cases}
cvr_{ij_pk},  & \text{if $k \in C_2$} \\
0, & \text{else}
\end{cases}
\end{split}
\end{equation}

Then we can calculate the value of $(rpm_{ij_pk} - \alpha_kcost_{ij_pk}+\gamma I_{C_1}(k) + \delta I_{C_2}(k))$ of each ad in the bidding landscape $L_i$. We sort the ads in $L_i$ by this score and choose the top $P$ ads to construct the best slate (while maintaining the ordering in $L_i$). The time complexity is reduced to $\mathcal{O}(P_i\log(P_i))$. It's worth noting that, all these prediction values like $rpm_{ij_pk}$ are calculated based on the order of the original landscape so that there might be a slight inaccuracy by the influence of position bias and GSP mechanism mentioned in subsection 4.1.

\subsection{Offline Simulation System}
In this work, we need to get the landscape without budget constraints for training and approximate the results by different approaches to measure their performances. We build an offline replay system based on the real ad serving system to obtain training data. Compared with applying to real online system, build a replay system can break out budget constraints and get all the candidate ads. 

Our replay system records each request with the traffic information as well as auction information including estimated ctr, cvr and click price of each candidate. The raw data is organized as a table partitioned by time, which amounts to dozens of TBs each day. Based on the replay data from real traffic, we can apply any allocation method on the simulation system to obtain the ads displayed in the page.

\begin{figure}[htb]
\floatconts
  {fig:replay2}
  {\caption{the click and cost gap between simulation system and online system with different ranking and pricing schemas}}
  {
    \subfigure[][c]{\label{fig:replay:pos_bias_click} \includegraphics[width=0.4\linewidth]{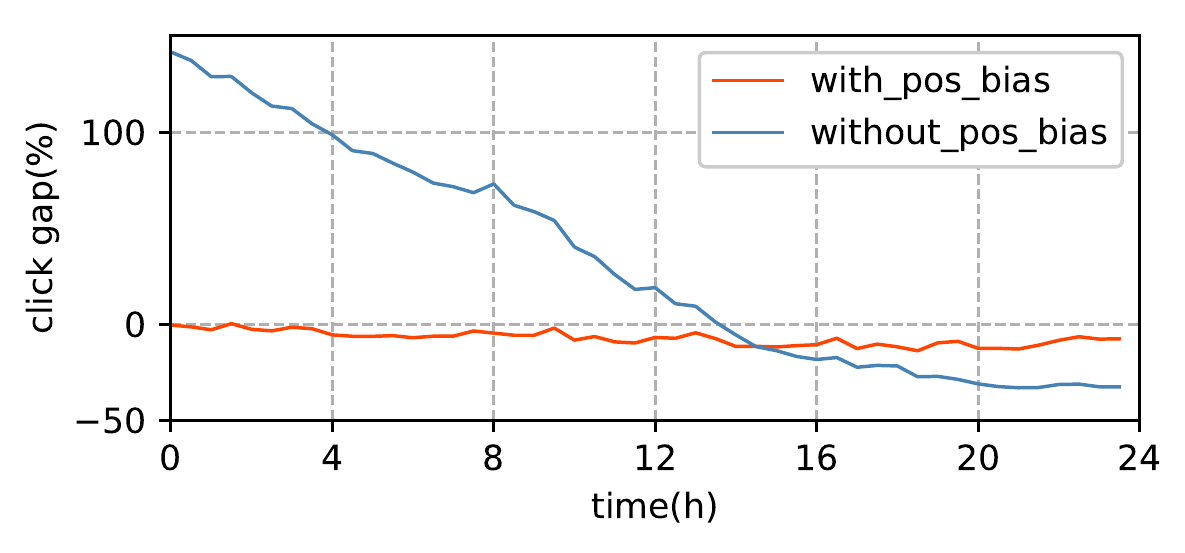}}%
    \qquad
    \subfigure[][c]{\label{fig:replay:bid_price_cost} \includegraphics[width=0.4\linewidth]{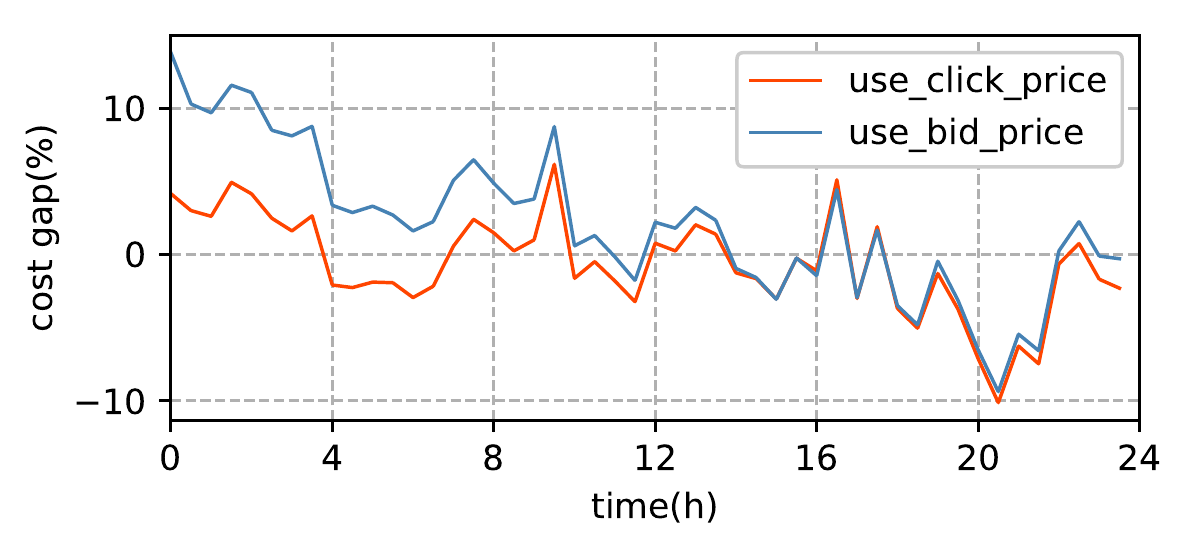}}
   }
\end{figure}

Furthermore, we try to make the results of online ad serving system and offline simulation system more consistent by improving ranking and pricing schemas. Fig.~\ref{fig:replay:pos_bias_click} demonstrates the click gaps between online and offline system. It shows that the result with considering position bias dramatically outperforms better than that without position bias. Additionally, we calculate real click price by GSP instead of using original bidprice, which makes obvious improvement shown by time in Fig.~\ref{fig:replay:bid_price_cost}.

\section{Experiments}
\subsection{Setup}
In this section, we conduct some comprehensive experiments to validate the effectiveness of the proposed algorithm. It's worth noting that our approach can be applied to both displaying and search advertising. Here we focus on one of our search advertising scenes and run experiments on the simulation system using real data from Alibaba.com. We choose two days data in May 2019 for training and testing respectively. The dataset covers nearly one hundred thousand campaigns and request data that contains tens of millions of records with auction information such as predicted $ctr$, $cvr$, \textit{click price} and other detail information.

In this paper, we demonstrate two types of experiments and analyze their results:
\begin{itemize}
	\item {\textbf{Single-Objective(SO)}}: Single-Objective optimization with budget constraints is a specific case of our model. We run the experiments of click optimization which is a common performance goal in E-commerce scenario, and compare the results with previous work \cite{karande2013optimizing}.
	\item {\textbf{Multi-Objective(MO)}}: We try to maximize the platform revenue and improve different performance goals simultaneously. We randomly choose 70\% of campaigns to improve their clicks, and 30\% of campaigns to increase their conversions. We leverage $T_{cy}$ and $T_{vy}$ as a controller and search the parameter space to get different results of optimal solutions. 
\end{itemize}

We also introduce two different approaches as follows:
\begin{itemize}
	\item {\textbf{Greedy Heuristic Policy (GHP)}}: Let advertisers participate in auctions as many as possible until they hit their budget and then make them ineligible for the rest of the day. All the relative improvement performance are calculated relative to the \textbf{baseline} of the \textbf{GHP}.
	\item {\textbf{Optimized Throttling (OT)}}: Given the objective, we rank all the requests for a campaign by its' performance goal, and choose the request with the top values until the budget is exhausted. They used the performance value of the last request chosen as a threshold to filter the real-time request. The algorithm detail is introduced in \cite{karande2013optimizing}.
\end{itemize}

\subsection{Results of Single-Objective}
In this subsection, we discuss the experimental results of click and conversion optimization with budget constraints respectively. The results are compared with \textbf{OT} Algorithm in Table 1, and the improvements are all calculated relative to the baseline of \textbf{GHP}.

The results in \tableref{tab:click_compare} reveal that our algorithm \textbf{SO-clk} dramatically outperforms \textbf{GHP} and \textbf{OT-clk} in CLK. Compared with \textbf{OT-clk}, our approach is also better in REV. As mentioned in Section 2.0, \textbf{OT} doesn't work well in a complex competition environment in E-commerce advertising and may cause unnecessary competitions. Besides the overall results, the performance of each campaign is also what we concerned. CLK+ represents the ratio of campaigns whose click number is increased while CLK- is the ratio of campaigns whose click number is declined.

\tableref{tab:conversion_compare} shows the results of conversion optimization. \textbf{SO-cvn} outperforms \textbf{GHP} and \textbf{OT-cvn} in both REV and CNV. And more campaigns get better performance in CVN by \textbf{SO-cvn} than \textbf{OT-cvn} and \textbf{GHP}.

\begin{table}[htbp]
	\centering
	\caption{Relative performance and ratio of campaigns compared with OT-clk}
	\begin{tabular}{ccccccc}
		\toprule
		Algorithm   & $\Delta_{CLK}$   & $\Delta_{REV}$ & $CLK_{+}$ & $CLK_{-}$ \\
		\midrule
		\midrule
		GHP & - & - & - & - \\
		OT-clk & 7.08\% & -5.93\% & 32.29\% & 46.53\% \\
		SO-clk & \textbf{28.63}\% & -4.95\% & \textbf{56.61}\% & 24.71\% \\
		\bottomrule
	\end{tabular}%
\label{tab:click_compare}
\end{table}

\begin{table}[!htb]
	\centering
	\caption{Relative performance and ratio of campaigns compared with OT-cvn}
	\begin{tabular}{ccccccc}
		\toprule
		Algorithm   & $\Delta_{CVN}$  & $\Delta_{REV}$ & $CVN_{+}$ & $CVN_{-}$  \\
		\midrule
		\midrule
		GHP & - & - & - & - \\
		OT-cvn & 13.71\% & -7.66\% & 26.5\% & 52.4\%  \\
		SO-cvn & \textbf{39.65}\% & -5.85\% & \textbf{46.33}\% & 34.98\% \\
		\bottomrule
	\end{tabular}%
\label{tab:conversion_compare}
\end{table}%
\begin{table}[!htb]
	\centering
	\caption{Relative performance compared with OT-clk-cvn}
	\begin{tabular}{ccccccc}
		\toprule
		Algorithm & $\Delta_{REV}$ & $\Delta_{CLK}$ & $\Delta_{CLK_{C1}}$ & $\Delta_{CVN}$ & $\Delta_{CVN_{C2}}$\\
		\midrule
		\midrule
		GHP & - & - & - & - & - \\
		OT-clk-cvn & -5.46\% & 5.80\% & 10.20\% & 9.78\% & 26.6\% \\
		MO-$T_{cy}15\%,T_{vy}30\%$ & \textbf{5.61}\% & 12.55\% & \textbf{13.91}\% &18.00\% & \textbf{28.25}\% \\
		\bottomrule
	\end{tabular}%
	\label{tab:mo_clk_cvn_compare}%
\end{table}%

\begin{table}[!htb]
	\centering
	\caption{The ratio of campaigns }
	\begin{tabular}{ccccc}
		\toprule
		Campaign Ratio & $CLK_{+}$ & $CLK_{-}$ & $CVN_{+}$ & $CVN_{-}$  \\
		\midrule
		\midrule
		GHP & - & - & - & -  \\
		OT-clk-cvn & 46.13\% & 32.68\% & 43.69\% & 34.92\%  \\
		MO-$T_{cy}15\%,T_{vy}30\%$ & \textbf{60.46}\% & 19.80\% & \textbf{64.08}\% & 15.89\%  \\
		\bottomrule
	\end{tabular}%
	\label{tab:mo_clk_cvn_camps_compare}%
\end{table}%

\subsection{Results of Multi-Objective}
Here we try to maximize the overall revenue and improve advertisers' performance goals (clicks and conversions) simultaneously. We randomly choose 70\% of campaigns ($C1$) to improve their clicks, and 30\% of campaigns ($C2$) to increase their conversions. Besides, we leverage $T_{cy}$ (the click constraint of $C1$)  and $T_{vy}$ (the conversion constraint of $C2$) as a controller and search the parameter space to get different results of optimal solutions. Our algorithm is compared with \textbf{OT}, and the results relative to \textbf{GHP} are summarized in \tableref{tab:mo_clk_cvn_compare} and \tableref{tab:mo_clk_cvn_camps_compare}.

$\Delta_{CLK}$ in \tableref{tab:mo_clk_cvn_compare} represents the increase of total clicks of all the campaigns relative to \textbf{GHP}, and $\Delta_{CLK_{C1}}$ refers to click increment of the campaigns with click goal (70\% of campaigns trying to improve their clicks). Likewise, $\Delta_{CVN}$ is the total conversion increment while $\Delta_{CVN_{C2}}$ is the result of the campaigns with conversion goal. Here $T_{cy}15\%$ in \tableref{tab:mo_clk_cvn_compare} stands for the 15\% percentage of improvement on the click of the baseline \textbf{GHP} with $C1$ in the train data. And $T_{vy}30\%$ represents the 30\% percentage of improvement on the conversions of the baseline \textbf{GHP} with $C2$. Apparently, our algorithm gets better performance than \textbf{OT-clk-cvn} and \textbf{GHP} not only in the overall campaigns but also the target campaigns since our algorithm optimizes all requests throughout the day while \textbf{OT} only applies to the head queries. Although our model can improve the overall performance, it can not guarantee that the performance of every campaign is improved. The results in \tableref{tab:mo_clk_cvn_camps_compare} show that most campaigns with performance goals gets better.

Furthermore, we split the data into 48 time periods (30 minutes between two consecutive time intervals) and study the result of different algorithms on the measures in consecutive time periods in Fig.~\ref{fig:clk_cvn_rev}. The curves show that our algorithm sacrifice the revenue in exchange for the click and conversion gain at the beginning of the day. However the overall performances of our method get better than \textbf{OT} and \textbf{GHP}.

\begin{figure}[htb]
\floatconts
  {fig:clk_cvn_rev}
  {\caption{CLK, CVN and REV over time with different algorithms}}
  {
    \subfigure[CLK][c]{\includegraphics[width=0.33\linewidth]{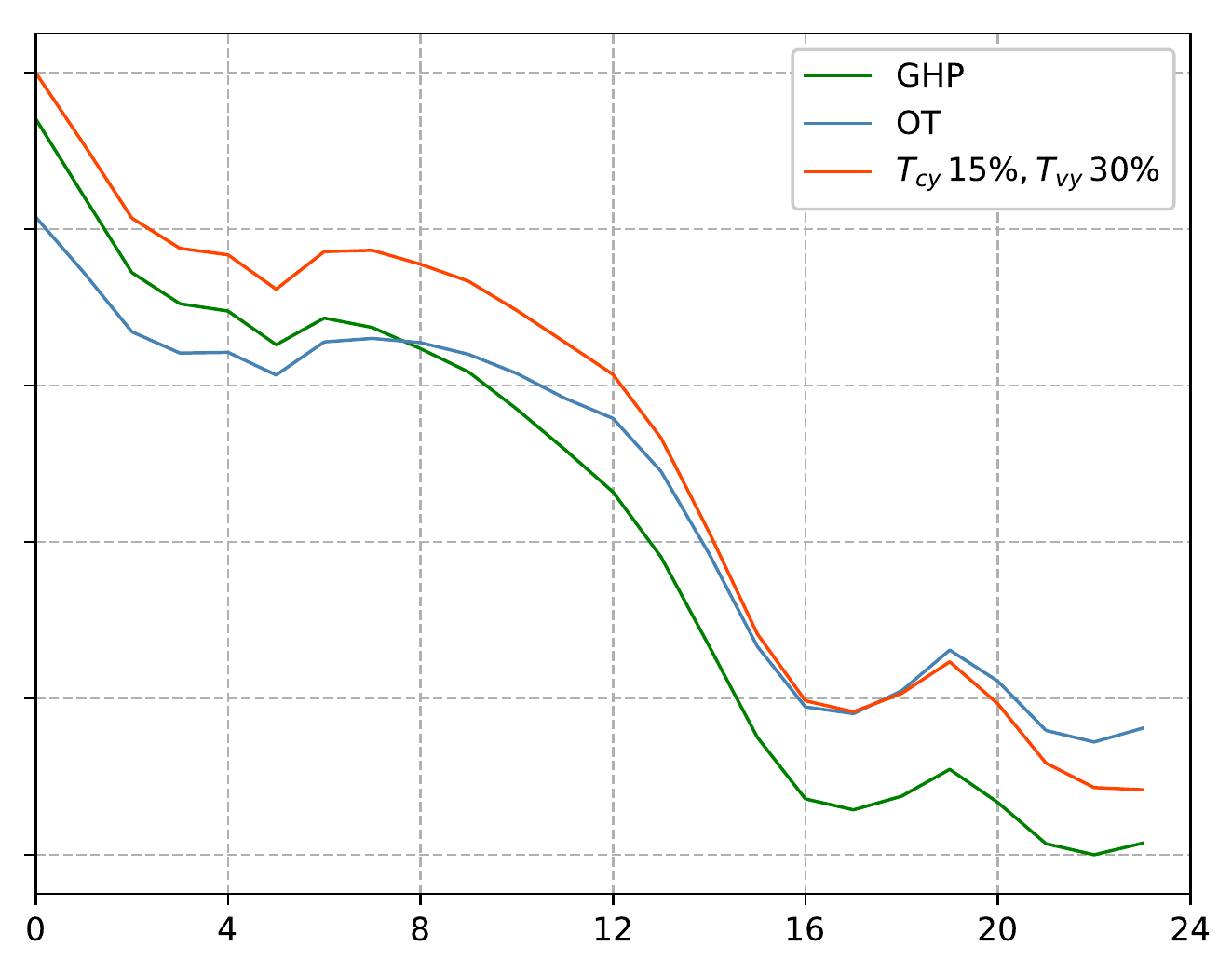}\label{rpm_indic:CLK}\hspace{1pt}}
    \qquad
    \subfigure[CVN][c]{\includegraphics[width=0.33\linewidth]{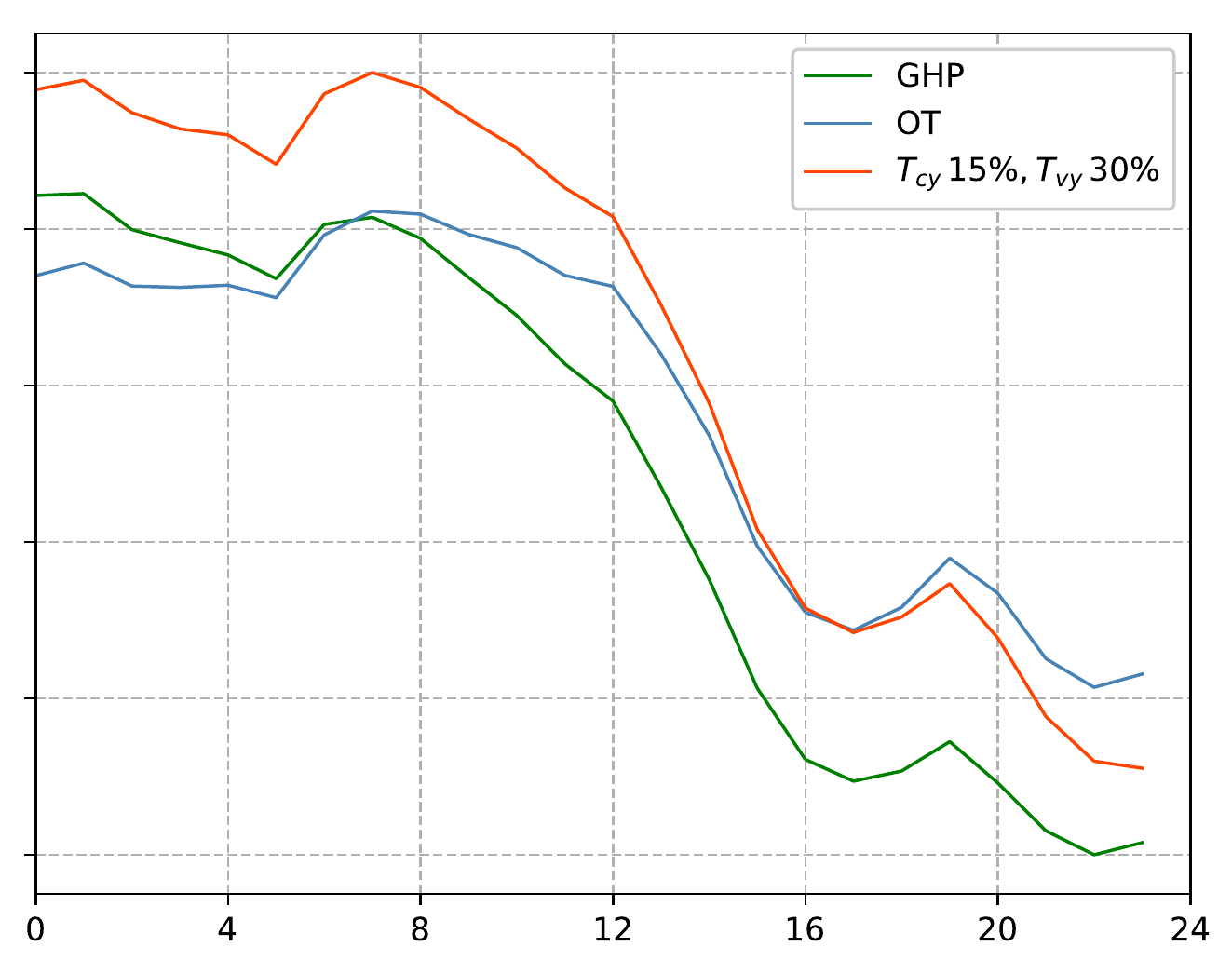}\label{rpm_indic:CVN}\hspace{1pt}}
    \qquad
    \subfigure[REV][c]{\includegraphics[width=0.33\linewidth]{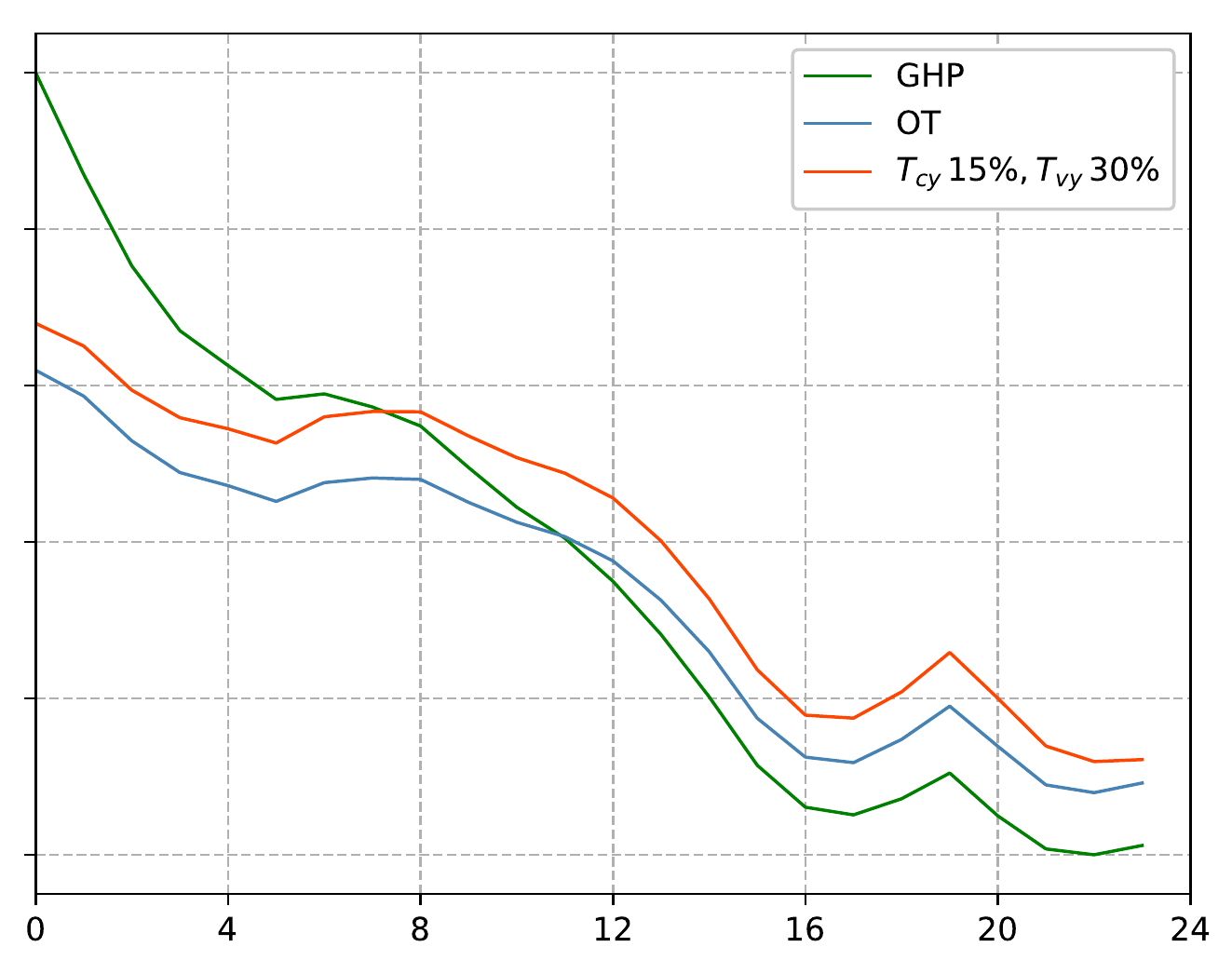}\label{rpm_indic:REV}\hspace{1pt}}
   }
\end{figure}

\section{Conclusion}
In this paper, we focus on optimizing budget constrained spend in E-commerce performance-based advertising. The objective of advertisers is usually to spend out the budget to maximize various performance goals while the ad serving system aims to optimize revenue on behalf of the platform. We propose an approach based on linear programming and build a simulation system to solve the problem. We run large-scale real data experiments in Alibaba.com to demonstrate the benefit of using this algorithm. The results show that it can significantly achieve delivery goals and improve the overall revenue of the platform.

Our future work will mainly focus on incorporating real-time data into the problem setup to improve accuracy. In addition, the offline simulation system needs to be improved.

\acks{We would like to thank the Alibaba.com Ads Engineering team for helpful engineering support.}

\bibliography{optimal_delivery_orsum19}

\begin{thebibliography}{19}
\providecommand{\natexlab}[1]{#1}
\providecommand{\url}[1]{\texttt{#1}}
\expandafter\ifx\csname urlstyle\endcsname\relax
  \providecommand{\doi}[1]{doi: #1}\else
  \providecommand{\doi}{doi: \begingroup \urlstyle{rm}\Url}\fi

\bibitem[Abrams et~al.(2007)Abrams, Mendelevitch, and
  Tomlin]{abrams2007optimal}
Zoe Abrams, Ofer Mendelevitch, and John Tomlin.
\newblock Optimal delivery of sponsored search advertisements subject to budget
  constraints.
\newblock In \emph{Proceedings of the 8th ACM conference on Electronic
  commerce}, pages 272--278. ACM, 2007.

\bibitem[Agarwal et~al.(2014)Agarwal, Ghosh, Wei, and You]{agarwal2014budget}
Deepak Agarwal, Souvik Ghosh, Kai Wei, and Siyu You.
\newblock Budget pacing for targeted online advertisements at linkedin.
\newblock In \emph{Proceedings of the 20th ACM SIGKDD international conference
  on Knowledge discovery and data mining}, pages 1613--1619. ACM, 2014.

\bibitem[Bhalgat et~al.(2012)Bhalgat, Feldman, and Mirrokni]{bhalgat2012online}
Anand Bhalgat, Jon Feldman, and Vahab Mirrokni.
\newblock Online allocation of display ads with smooth delivery.
\newblock In \emph{Proceedings of the 18th ACM SIGKDD international conference
  on Knowledge discovery and data mining}, pages 1213--1221. ACM, 2012.

\bibitem[Chapelle and Zhang(2009)]{chapelle2009dynamic}
Olivier Chapelle and Ya~Zhang.
\newblock A dynamic bayesian network click model for web search ranking.
\newblock In \emph{Proceedings of the 18th international conference on World
  wide web}, pages 1--10. ACM, 2009.

\bibitem[Chen et~al.(2012)Chen, Ma, Mandalapu, Nagarjan, Shanmugasundaram,
  Vassilvitskii, Vee, Yu, and Zien]{chen2012ad}
Peiji Chen, Wenjing Ma, Srinath Mandalapu, Chandrashekhar Nagarjan, Jayavel
  Shanmugasundaram, Sergei Vassilvitskii, Erik Vee, Manfai Yu, and Jason Zien.
\newblock Ad serving using a compact allocation plan.
\newblock In \emph{Proceedings of the 13th ACM Conference on Electronic
  Commerce}, pages 319--336. ACM, 2012.

\bibitem[Chen et~al.(2011)Chen, Berkhin, Anderson, and Devanur]{chen2011real}
Ye~Chen, Pavel Berkhin, Bo~Anderson, and Nikhil~R Devanur.
\newblock Real-time bidding algorithms for performance-based display ad
  allocation.
\newblock In \emph{Proceedings of the 17th ACM SIGKDD international conference
  on Knowledge discovery and data mining}, pages 1307--1315. ACM, 2011.

\bibitem[Chervonenkis et~al.(2013)Chervonenkis, Sorokina, and
  Topinsky]{chervonenkis2013optimization}
Alexey Chervonenkis, Anna Sorokina, and Valery~A Topinsky.
\newblock Optimization of ads allocation in sponsored search.
\newblock In \emph{Proceedings of the 22nd International Conference on World
  Wide Web}, pages 121--122. ACM, 2013.

\bibitem[Craswell et~al.(2008)Craswell, Zoeter, Taylor, and
  Ramsey]{craswell2008experimental}
Nick Craswell, Onno Zoeter, Michael Taylor, and Bill Ramsey.
\newblock An experimental comparison of click position-bias models.
\newblock In \emph{Proceedings of the 2008 international conference on web
  search and data mining}, pages 87--94. ACM, 2008.

\bibitem[Devanur et~al.(2011)Devanur, Jain, Sivan, and
  Wilkens]{devanur2011near}
Nikhil~R Devanur, Kamal Jain, Balasubramanian Sivan, and Christopher~A Wilkens.
\newblock Near optimal online algorithms and fast approximation algorithms for
  resource allocation problems.
\newblock In \emph{Proceedings of the 12th ACM conference on Electronic
  commerce}, pages 29--38. ACM, 2011.

\bibitem[Dupret and Piwowarski(2008)]{dupret2008user}
Georges~E Dupret and Benjamin Piwowarski.
\newblock A user browsing model to predict search engine click data from past
  observations.
\newblock In \emph{Proceedings of the 31st annual international ACM SIGIR
  conference on Research and development in information retrieval}, pages
  331--338. ACM, 2008.

\bibitem[Edelman et~al.(2007)Edelman, Ostrovsky, and
  Schwarz]{edelman2007internet}
Benjamin Edelman, Michael Ostrovsky, and Michael Schwarz.
\newblock Internet advertising and the generalized second-price auction:
  Selling billions of dollars worth of keywords.
\newblock \emph{American economic review}, 97\penalty0 (1):\penalty0 242--259,
  2007.

\bibitem[He et~al.(2013)He, Chen, Wang, and Liu]{he2013game}
Di~He, Wei Chen, Liwei Wang, and Tie-Yan Liu.
\newblock A game-theoretic machine learning approach for revenue maximization
  in sponsored search.
\newblock In \emph{IJCAI}, pages 206--212, 2013.

\bibitem[Karande et~al.(2013)Karande, Mehta, and
  Srikant]{karande2013optimizing}
Chinmay Karande, Aranyak Mehta, and Ramakrishnan Srikant.
\newblock Optimizing budget constrained spend in search advertising.
\newblock In \emph{Proceedings of the sixth ACM international conference on Web
  search and data mining}, pages 697--706. ACM, 2013.

\bibitem[Lee et~al.(2013)Lee, Jalali, and Dasdan]{lee2013real}
Kuang-Chih Lee, Ali Jalali, and Ali Dasdan.
\newblock Real time bid optimization with smooth budget delivery in online
  advertising.
\newblock In \emph{Proceedings of the Seventh International Workshop on Data
  Mining for Online Advertising}, page~1. ACM, 2013.

\bibitem[Mehta et~al.(2013)]{mehta2013online}
Aranyak Mehta et~al.
\newblock Online matching and ad allocation.
\newblock \emph{Foundations and Trends{\textregistered} in Theoretical Computer
  Science}, 8\penalty0 (4):\penalty0 265--368, 2013.

\bibitem[Xu et~al.(2015)Xu, Lee, Li, Qi, and Lu]{xu2015smart}
Jian Xu, Kuang-chih Lee, Wentong Li, Hang Qi, and Quan Lu.
\newblock Smart pacing for effective online ad campaign optimization.
\newblock In \emph{Proceedings of the 21th ACM SIGKDD International Conference
  on Knowledge Discovery and Data Mining}, pages 2217--2226. ACM, 2015.

\bibitem[Zhang et~al.(2014)Zhang, Yuan, and Wang]{zhang2014optimal}
Weinan Zhang, Shuai Yuan, and Jun Wang.
\newblock Optimal real-time bidding for display advertising.
\newblock In \emph{Proceedings of the 20th ACM SIGKDD international conference
  on Knowledge discovery and data mining}, pages 1077--1086. ACM, 2014.

\bibitem[Zhang et~al.(2018)Zhang, Wei, Meng, Hu, and Wang]{zhang2018whole}
Weiru Zhang, Chao Wei, Xiaonan Meng, Yi~Hu, and Hao Wang.
\newblock The whole-page optimization via dynamic ad allocation.
\newblock In \emph{Companion of the The Web Conference 2018 on The Web
  Conference 2018}, pages 1407--1411. International World Wide Web Conferences
  Steering Committee, 2018.

\bibitem[Zhu et~al.(2009)Zhu, Wang, Yang, Wang, Yan, and Chen]{zhu2009revenue}
Yunzhang Zhu, Gang Wang, Junli Yang, Dakan Wang, Jun Yan, and Zheng Chen.
\newblock Revenue optimization with relevance constraint in sponsored search.
\newblock In \emph{Proceedings of the Third International Workshop on Data
  Mining and Audience Intelligence for Advertising}, pages 55--60. ACM, 2009.

\end{thebibliography}

\appendix

\end{document}